\documentclass[10pt,twocolumn,letterpaper]{article}

\usepackage{cvpr}
\usepackage{graphicx}
\usepackage{amsmath}
\usepackage{amssymb}
\usepackage{booktabs}
\usepackage{graphicx}
\usepackage{amsmath}
\usepackage{amssymb}
\usepackage{booktabs,subcaption,amsfonts,dcolumn}
\usepackage{dsfont}
\usepackage{times}
\usepackage{epsfig}
\usepackage{color, colortbl}
\usepackage{multirow}
\usepackage{enumitem}
\usepackage{amsmath}
\usepackage{makecell}
\usepackage{hhline}
\usepackage{cellspace}
\usepackage{bbding}
\usepackage{bm}
\usepackage{nicefrac}     
\usepackage{microtype}     
\usepackage{verbatim}
\usepackage{color}
\usepackage{float}
\usepackage{enumitem}
\usepackage{tabulary,overpic,xcolor}
\definecolor{mygray}{gray}{.92}
\definecolor{myred}{rgb}{1,0.5,0.5}
\definecolor{myred2}{rgb}{0.75,0,0}
\usepackage{pifont}

\newcommand{\myparagraph}[1]{{\vspace{0.5em} \noindent \bf #1}}
\newlength\savewidth\newcommand\shline{\noalign{\global\savewidth\arrayrulewidth
		\global\arrayrulewidth 1pt}\hline\noalign{\global\arrayrulewidth\savewidth}}
\newcolumntype{x}[1]{>{\centering\arraybackslash}p{#1pt}}

\usepackage[utf8]{inputenc}
\newcommand{\authorskip}{\hspace{12mm}}
\usepackage{caption}
\usepackage{cuted}
\newcommand\blfootnote[1]{
  \begingroup
  \renewcommand\thefootnote{}\footnote{#1}
  \addtocounter{footnote}{-1}
  \endgroup
}
\usepackage[pagebackref,breaklinks,colorlinks]{hyperref}

\usepackage[capitalize]{cleveref}
\crefname{section}{Sec.}{Secs.}
\Crefname{section}{Section}{Sections}
\Crefname{table}{Table}{Tables}
\crefname{table}{Tab.}{Tabs.}

\begin{document}

\title{GRiT: A Generative Region-to-text Transformer for Object Understanding}

\author{
 Jialian Wu$^{1\dagger}$ \authorskip Jianfeng Wang$^{2}$ \authorskip  Zhengyuan Yang$^{2}$ \authorskip
 Zhe Gan$^{2}$ \\ Zicheng Liu$^{2}$ \authorskip Junsong Yuan$^{1}$ \authorskip Lijuan Wang$^{2}$\\[3mm]
 $^1$State University of New York at Buffalo ~~~~~~~~~~~~~~ $^2$Microsoft \\
 {\tt\small \{jialianw,jsyuan\}@buffalo.edu}\\ {\tt\small\{jianfw,zhengyang,zhe.gan,zliu,lijuanw\}@microsoft.com}
}

\twocolumn[{
\vspace{-1em}
\maketitle
\vspace{-1em}

\begin{center}
    \centering 
    \includegraphics[width=1\linewidth]{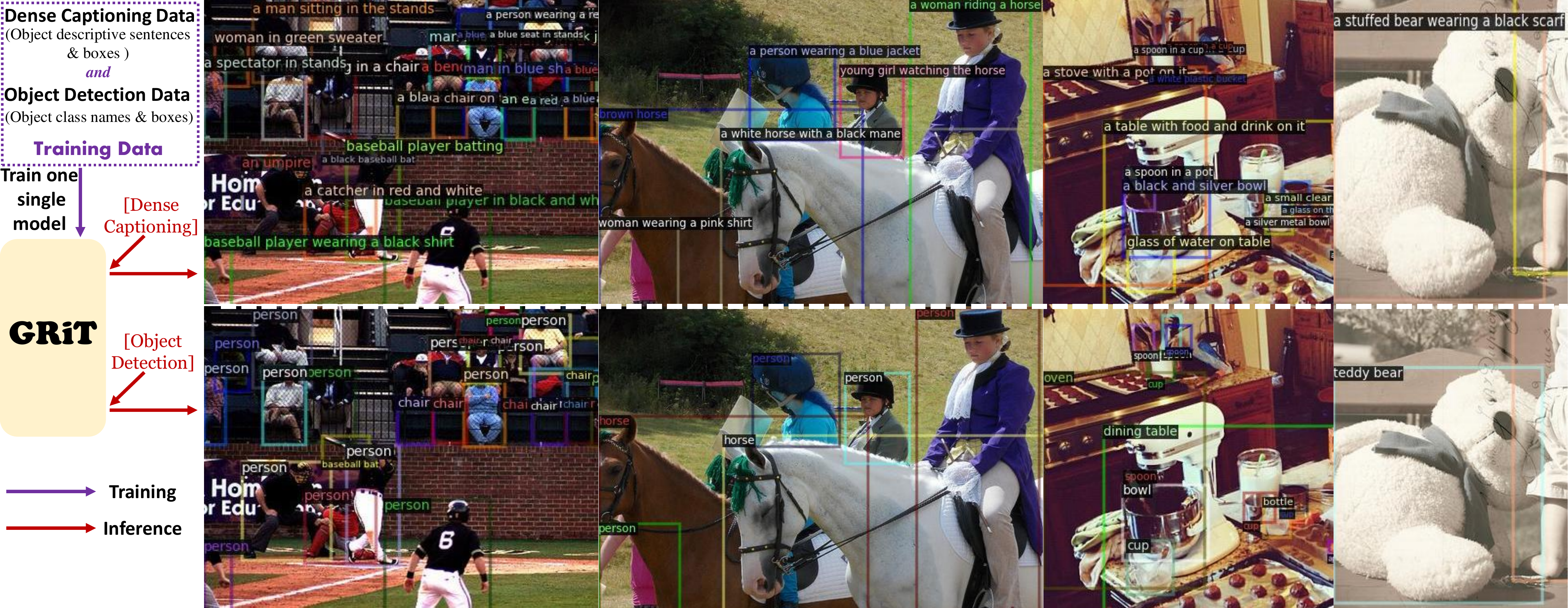}
    \vspace{-6mm}
    \captionof{figure}{\textbf{GRiT's object understanding pipeline with multi-task training.} GRiT is a general object understanding framework that localizes objects and generates object description texts in \emph{free-form}. The right figures show the predictions from the model trained on object detection data and dense (object) captioning data together. During inference, the trained GRiT can generate either simple class names for object detection task, or rich descriptive sentences for dense captioning task, instructed by two special tokens \texttt{[ObjectDet]} and \texttt{[DenseCap]}, respectively. Zoom in for the best viewing.}
    \label{fig:fig1}
\end{center}
}] 

\begin{abstract}
\vspace{-3mm}
\blfootnote{$^\dagger$Work was done when the author interned at Microsoft.}
This paper presents a Generative RegIon-to-Text transformer, GRiT, for object understanding. The spirit of GRiT is to formulate object understanding as $<$region, text$>$ pairs, where region locates objects and text describes objects. For example, the text in object detection denotes class names while that in dense captioning refers to descriptive sentences. Specifically, GRiT consists of a visual encoder to extract image features, a foreground object extractor to localize objects, and a text decoder to generate open-set object descriptions. With the same model architecture, GRiT can understand objects via not only simple nouns, but also rich descriptive sentences including object attributes or actions. Experimentally, we apply GRiT to object detection and dense captioning tasks. GRiT achieves competitive detection accuracy (60.4 AP on COCO 2017 test-dev) and new state-of-the-art dense captioning performance (15.5 mAP on Visual Genome). Code is available at \url{https://github.com/JialianW/GRiT}
\end{abstract}

\section{Introduction}
\label{sec:intro}
\begin{figure}[t]
	\centering
	\includegraphics[width=1\linewidth]{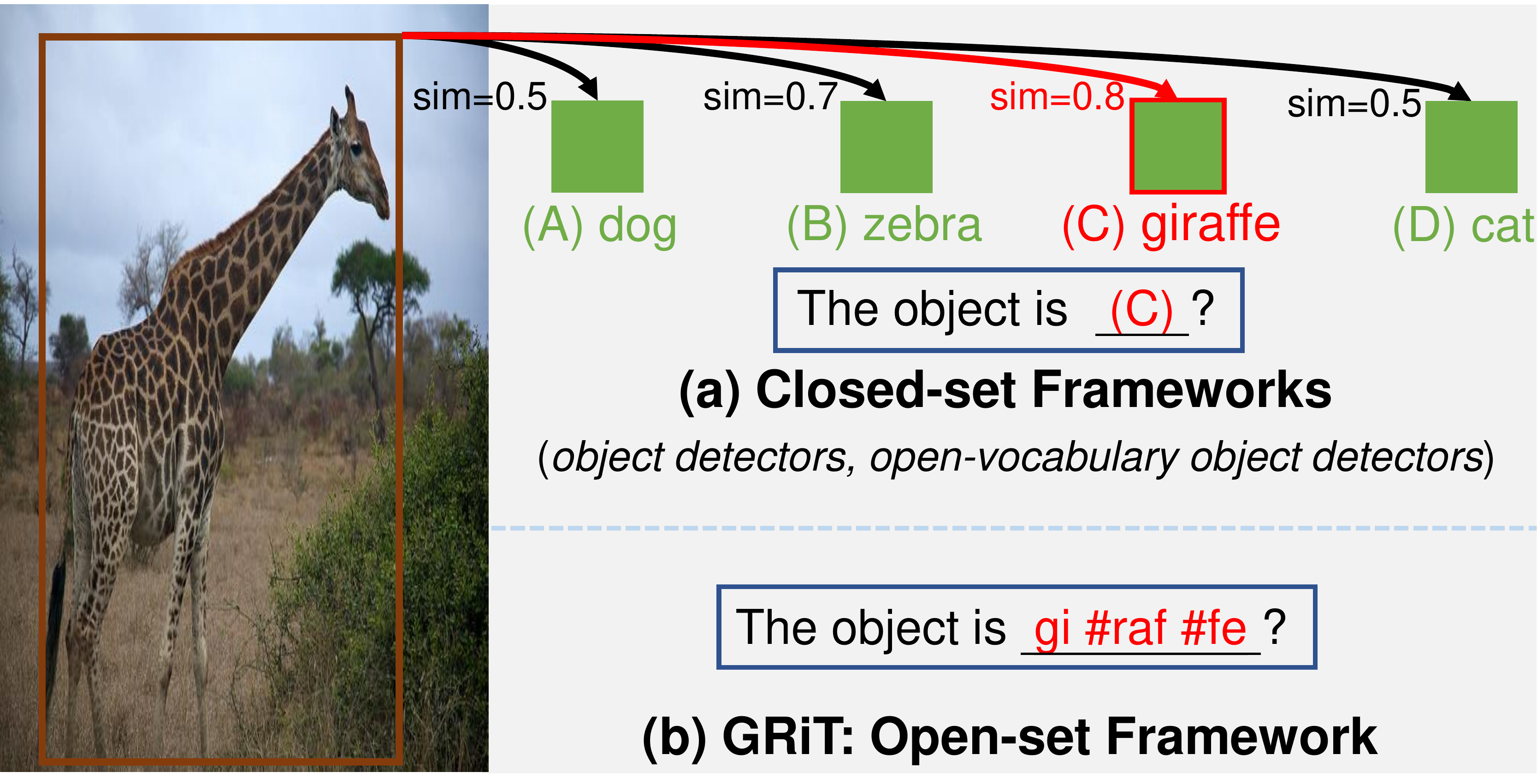}
	\caption{\textbf{GRiT \emph{vs.} Closed-set object recognition models.} Closed-set approaches classify objects among a predefined set of categories by choosing the class embedding that has the maximum similarity to the given object feature. In contrast, GRiT directly generates text tokens to spell the words that describe the object. In this example, ``\texttt{giraffe}'' is spelled by three text tokens generated by GRiT.}
	\label{fig:fig3}
\end{figure}

Great efforts have been made in object detection task~\cite{ren2015faster,zhou2019objects,cai2018cascade,li2022exploring}. To answer `what the object is?', object detection models classify objects among a \emph{closed-set} of object classes, where class names are usually simple nouns. Recent open-vocabulary object detectors~\cite{zhong2022regionclip,zhou2022detecting,zareian2021open,gu2021open,gaoopen,li2022grounded} exploit fertile vision and language data to enable object detectors to recognize object classes that do not exist in object detection data. Open-vocabulary object detectors, however, still require to define a closed-set of class name embeddings to achieve object classification. The ``open-vocabulary'' only means some of the class name embeddings are not associated with object detection data during training. As illustrated in Fig.~\ref{fig:fig3} (a), these closed-set frameworks behave like performing a multiple-choice question, choosing the most likely answer from a limited number of candidates.

In contrast, humans understand objects in an \emph{open-set} configuration that does not limit the number of object categories and can therefore learn new objects effortlessly. Also, humans perceive auxiliary information associated with the object to improve understanding, \eg, color, shape, and action expressed through adjectives and verbs. Toward human-like object understanding, we propose a Generative RegIon-to-Text transformer, coined as GRiT, which requires no predefined list of category names and can understand object in a more descriptive fashion. 

Given an input image, GRiT localizes all presented objects and generates object descriptions for each of them in \emph{free-form}. Specifically, GRiT consists of three main components: a visual encoder, a foreground object extractor, and a text decoder. The visual encoder extracts image features, on which the foreground object extractor detects foreground object regions and crops object features. Taking object features as input, the text decoder autoregressively generates a list of text tokens to describe the given object, where each word is tokenized by one or multiple text tokens (\emph{a.k.a.} sub-words) by the WordPiece~\cite{wu2016google, schuster2012japanese} model. In addition to simple noun category names (\eg, cat, giraffe), GRiT can also generate rich descriptive sentences, including but not limited to object attributes and actions as shown in Fig.~\ref{fig:fig1}. In this way, GRiT achieves general object understanding that can unify various region-level tasks into a single framework, \eg, object detection and dense (object) captioning.

With the same architecture, GRiT can train on the short-description task (\eg, object detection) and the long-description task (\eg, dense captioning) separately or together. Joint training on short- and long-description tasks without any adaptations can confuse the model to generate descriptions for the correct task. To solve this issue, we add a special token \texttt{[task]} to control GRiT's text decoder to predict task-specific object descriptions. 

GRiT is an \emph{open-set} object understanding framework that can describe objects with unlimited words, as any word can be represented by a combination of text tokens. As shown in Fig.~\ref{fig:fig3} (b), GRiT's \emph{open-set} way is similar to performing a fill-in-the-blank question that spells out the answer from scratch. This is more universal but much more challenging compared to the closed-set object recognition frameworks such as object detectors and open-vocabulary object detectors. Such an open-set nature is friendly as data grow and evolve. GRiT can continuously learn newly added object classes without adapting model architecture.

In summary, this paper presents a general and open-set object understanding method that can unify multiple region-level tasks into a single framework. GRiT achieves 60.4 AP on COCO object detection~\cite{lin2014microsoft}. It is comparable to the closed-set standard object detectors, which is remarkable in light of our more challenging open-set way. On Visual Genome dense captioning~\cite{krishna2017visual}, GRiT obtains 15.5 mAP which surpasses all dense captioning models. We hope our work can inspire more future works on this general open-set object understanding.

\begin{figure*}[h]
	\centering
	\includegraphics[width=1\linewidth]{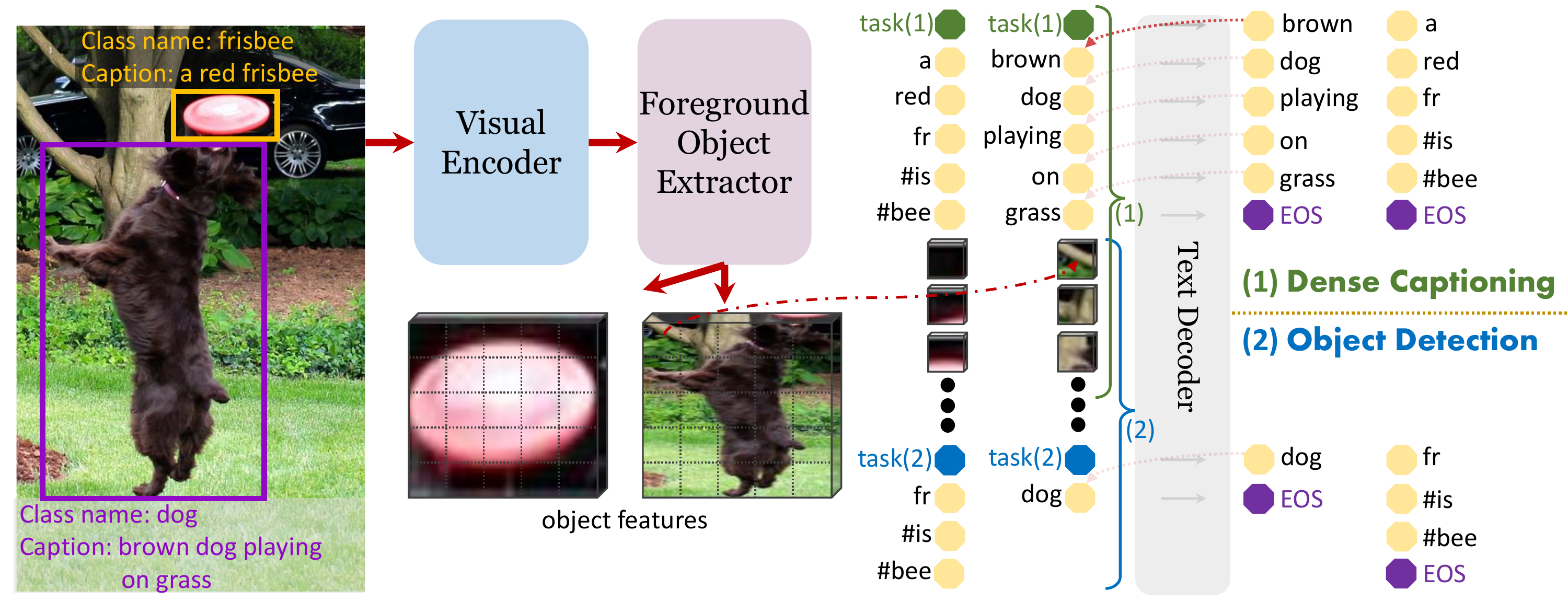}
	\caption{\textbf{Overview of GRiT.} Given an input image, the visual encoder extracts image features, from which the foreground object extractor predicts object boxes. Object features are derived by cropping image features using object boxes. Taking object features as input, the text decoder autoregressively generates text tokens one-by-one in the task-$i$ style, instructed by a begin token \texttt{[task]}$_i$.}
	\label{fig:overview}
\end{figure*}

\section{Related Work}
Continuous progress has been made to push the boundaries of standard object detection, from the classic R-CNN family~\cite{ren2015faster,chen2019hybrid,cai2018cascade} to the recent transformer approaches~\cite{zhang2022dino,carion2020end,zhu2020deformable}. Despite excellent object localization and classification accuracy, these frameworks are designed to recognize a fixed set of object categories.

Recent open-vocabulary object detection~\cite{zhong2022regionclip,zhou2022detecting,zareian2021open,gu2021open,gaoopen,li2022grounded} breaks the fixed category limit. The goal is to recognize object categories that are not in object detection datasets, using the knowledge learned from large-scale vision and language data. For example, RegionCLIP~\cite{zhong2022regionclip} exploits CLIP~\cite{radford2021learning} trained on hundreds of millions of image-text data to match region features with text embeddings, where the knowledge in CLIP helps to annotate region-text pairs not labeled in object detection data. ViLD~\cite{gu2021open} makes use of the CLIP language embeddings and distills the CLIP vision knowledge into its own visual backbone. More straightforwardly, Detic~\cite{zhou2022detecting} treats the full image as a whole box and learns directly from the large-scale image labels. These open-vocabulary object detectors can vary the size of the category set and recognize more categories beyond object detection datasets. However, the category set still needs to be closed and predefined by humans in order to construct a contrastive matrix with the object regions and have the model choose a category from it to label the object. Besides, they do not show the ability to generate descriptive sentences. 

In this work, we are inspired by the generative image-to-text transformer~\cite{wang2022git,hu2022scaling,yu2022coca}. The generative methods produce free-form words/sentences to achieve various image understanding tasks like image captioning/question answering/classification. Our GRiT extends the spirit to region-level object understanding, aiming to generate object descriptions in free-form with not only class names but also rich descriptive sentences. GRiT inherits the open-set feature of the generative methods, which does not need humans to define a category list and \emph{just spells out} object descriptions on its own.

In contrast to dense captioning models~\cite{johnson2016densecap,yang2017dense,li2019learning,yin2019context,shao2022region}, GRiT is a general object understanding framework unifying both object detection and dense captioning. In architecture, GRiT is advanced by the simple yet successful generative image-to-text transformer. This enables our state-of-the-art dense captioning performance without the need for complex object relation and context modeling as in previous dense captioning models.

\section{GRiT}
\label{sec:grit}
\subsection{Architecture}
As illustrated in Fig.~\ref{fig:overview}, GRiT comprises three major components: a visual encoder, a foreground object extractor, and a text decoder. GRiT is end-to-end in both training and inference. We adopt the same network structure for the visual encoder backbone and text decoder as the image-to-text transformer (GIT~\cite{wang2022git} here in this paper).

\myparagraph{Visual Encoder.} Given an input image, a visual encoder is applied to obtain image features. Our visual encoder consists of a backbone network, and a feature pyramid that is proven helpful for object detection~\cite{li2022exploring, lin2017feature}. As GIT$_{B/L}$~\cite{wang2022git}, we use ViT~\cite{dosovitskiy2020image} as the backbone network for main experiments\footnote{In experiments, we also ablate the Coswin-H backbone~\cite{yuan2021florence} adopted by GIT~\cite{wang2022git}.}. Different from image VL tasks, object understanding favors high-resolution input images, which can consume huge GPU memory in training ViT's self-attention. Therefore, we divide ViT's feature maps into non-overlapped windows with the size of $14\times14$, and compute self-attention only within the windows following~\cite{li2022exploring}. A relative positional encoding is added during the window self-attention as in~\cite{liu2021swin}. To exchange information across the windows, four evenly selected ViT blocks keep the original ViT self-attention scheme which computes global self-attention across all positions on the feature map. ViT extracts image features throughout in a single scale without hierarchies, \eg, $\frac{1}{16}\times$ image size, which is however incompatible with vanilla FPN~\cite{lin2017feature}. To build feature pyramid on top of ViT, we follow the idea of simple feature pyramid~\cite{li2022exploring} that produces multi-scale features by up/down-sampling from the last feature map of ViT. In this way, we construct five scales of feature maps $\{\frac{1}{8},\frac{1}{16},\frac{1}{32},\frac{1}{64},\frac{1}{128}\}$ for our feature pyramid.

\myparagraph{Foreground Object Extractor.} On top of the feature pyramid, our foreground object extractor detects foreground objects with bounding boxes and objectness scores. The foreground object extractor borrows the architecture of two-stage object detectors~\cite{ren2015faster,cai2018cascade,zhou2021probabilistic}, comprised by a proposal generator and an RoI head. The proposal generator produces a large amount of proposal boxes. Then, the RoI head refines the box position and predicts an objectness score indicating the confidence that the box contains a foreground object. Thus, it is a binary classifier in the RoI head for foreground/background classification, which is different from the multi-category classifier in standard object detectors. Lastly, the foreground object extractor removes highly overlapped boxes by NMS and the boxes with low objectness scores.

\myparagraph{Text Decoder.} Our text decoder is the core part of GRiT for understanding and describing objects in free-form. The text decoder takes object features as input and generates text tokens to describe the given object. To derive object features, we use object boxes produced by the foreground object extractor to crop image features to a fixed size, \eg, $14\times14$, and then flatten them into 1D vectors. Object features encode not only object appearance but also image context thanks to the global self-attention blocks in ViT. This is important for dense captioning task where object descriptions may contain context descriptions in other image regions as well, as shown by the example of ``young girl watching the horse'' in Fig.~\ref{fig:fig1}. To convert words into text tokens, we use the BERT's WordPiece tokenizer~\cite{devlin2018bert,wu2016google,schuster2012japanese}, where any word can be represented by a combination of text tokens from the overall 30,522 tokens of vocabulary. For example, the class name ``\texttt{giraffe}'' in COCO is converted into three text tokens ``\texttt{gi}'', ``\texttt{\#raf}'', and ``\texttt{\#fe}''. 

The text decoder is achieved by a 6-layer transformer equipped with a begin token \texttt{[task]}. The text decoder produces object descriptions by generating text tokens one-by-one in an autoregressive way until an end token \texttt{[EOS]}. Following GIT~\cite{wang2022git}, in each step, object features and previously generated tokens, including the begin token, are concatenated as input to the transformer. Both object features and text tokens are embedded into the same dimensions before feeding into the transformer, where we also add a positional encoding to the text embeddings. A seq2seq attention mask is applied to make sure text tokens are attending to object features and only previous text tokens, and object features are only attending to themselves.  At the end of the transformer, a linear layer projects the text embeddings into 30,522 text token logits, and a softmax is applied afterward to yield the score for each text token. The text token with the highest score is kept. To predict the class name ``\texttt{giraffe}'', GRiT needs to consecutively generate the three tokens ``\texttt{gi}'', ``\texttt{\#raf}'', and ``\texttt{\#fe}''. Thanks to this flexible application of text tokens, GRiT achieves an \emph{open-set} object understanding that can describe whatever we provide in training. 

Different tasks may have varied styles of object descriptions. For example, object detection task interprets objects by short class names, while dense captioning task describes objects with rich descriptive sentences including object attributes, quantity, or actions. Jointly training them can confuse the model in inference, not knowing which style of object descriptions it should generate. To solve this issue, we define a set of begin tokens $\{$\texttt{[task]}$_i\}_{i=1}^{T}$ for jointly training tasks with different styles of object descriptions. $T$ is the number of different styles of tasks. In training, we select \texttt{[task]}$_i$ as the begin token when the object description annotation is from the task-$i$. In this way, during inference, \texttt{[task]}$_i$ can inform the trained model to generate descriptions in the task-$i$ style. 

\subsection{Training} The training loss of GRiT consists of two major parts $L=L_{o}+L_{t}$, where $L_{o}$ and $L_{t}$ are for the foreground object extractor and text decoder, respectively. $L_{o}$ is the same as the standard object detector loss that includes box losses and classification losses for both the proposal generator and RoI head. $L_{t}$ is achieved by the language modeling (LM) loss as follows:
\begin{equation}
    L_{t} = \frac{1}{N+1}\sum_{i=1}^{N+1}\text{CE}(y_i, p(y_i|o,y_{0,...,i-1})),
\end{equation}
where $p(y_i|o,y_{0,...,i-1})$ is the predicted score for the $i$-th text token given the object features $o$ and previously generated text tokens $y_{0,...,i-1}$. $N$ is the number of text tokens in the given object description. $y_{0}$ and $y_{N+1}$ are the begin token and end token, respectively. CE is the cross-entropy loss with a label smoothing of 0.1. Note that the text decoder loss $L_{t}$ is only imposed on foreground objects predicted by the foreground object extractor.

\subsection{Inference}
\myparagraph{Beam Search.} Standard object detectors may yield multiple object class labels for one box to improve performance. To enable a similar mechanism in GRiT, we employ a beam search algorithm in the text decoder, which is commonly used in image captioning. Specifically, we select the top $k$ text tokens in terms of their scores when generating the first token of object description in addition to the begin token, where $k$ is the beam size. The text decoder then continues to decode $k$ object descriptions following these $k$ text tokens. In experiments, we find $k=3$ is sufficient for object detection on COCO. We do not use beam search for dense captioning.

\myparagraph{Object Scoring.} GRiT rates object predictions by objectness score from the foreground object extractor and an object description score from the text decoder. Since an object description may contain multiple text tokens, its score is computed by averaging the scores of all text tokens. The final object confidence score is computed by multiplying the square roots of these two scores.

\section{Experiments}
To evaluate GRiT's general object understanding capability, we experiment on the COCO dataset~\cite{lin2014microsoft} for object detection task and the Visual Genome (VG) dataset~\cite{krishna2017visual} for dense (object) captioning task.

\myparagraph{COCO.} COCO contains 80 object classes and all class names are nouns, in which 15 class names consist of two words, \eg, ``\texttt{tennis racket}''. Each class name in COCO is encoded by 1$\sim$3 text tokens. We train on COCO 2017 train and evaluate on COCO 2017 val and 2017 test-dev. \textbf{Evaluation Metric:} Object detection performance is evaluated by COCO box AP and AR measured across IoU thresholds of 0.5 to 0.95.
As our approach imposes no hard constraint in generating the class names, we remove boxes whose class names are not in COCO during evaluation.

\myparagraph{Visual Genome.} We use VG v1.0 train set and test set for training and evaluation. Following the original paper~\cite{johnson2016densecap}, we pre-process VG data to discard object descriptions with more than 15 words and convert symbols into English words, \eg, $^{\circ} \rightarrow$ ``\texttt{degree}''. The pre-processing leaves 77,396 images for the train set and 5,000 images for the test set. There are $\sim$4 million annotated region descriptions with over 50,000 unique words in the train set. The annotations contain some typographical mistakes like ``\texttt{tranportation}'' $\rightarrow$ ``\texttt{transportation}'', which we don't perform further processing as GRiT can be fault-tolerant to some extent. Different from COCO object detection, object descriptions in VG have adjectives and verbs in addition to nouns, describing object attributes, actions, etc. \textbf{Evaluation Metric:} Similar to object detection metric, dense captioning measures an mAP across a range of thresholds for both localization and description accuracy, following~\cite{johnson2016densecap}. For localization, it uses box IoU thresholds of .3, .4, .5, .6, .7. For language description, a METEOR score~\cite{laviemeteor} with thresholds of 0, .05, .1, .15, .2, .25 is used. The mAP is averaged by the APs across all pairwise of these two types of thresholds.

\subsection{Implementation Details}
\myparagraph{Visual Encoder.} We employ ViT-B, ViT-L, and ViT-H~\cite{dosovitskiy2020image} as the backbone of the visual encoder unless otherwise specified. The input image patch size is $16\times16$. A layer-wise learning rate decay~\cite{li2022exploring} of 0.7/0.8/0.9 is set for ViT-B/L/H. For feature pyramid, the feature maps of $\{\frac{1}{8},\frac{1}{16},\frac{1}{32}\}$ scales are constructed in the way of simple feature pyramid~\cite{li2022exploring}. The feature maps of $\{\frac{1}{64},\frac{1}{128}\}$ scales are built by downsampling from that of $\frac{1}{32}$ scale following~\cite{zhou2021probabilistic}.

\myparagraph{Foreground Object Extractor.} We use CenterNet~\cite{zhou2019objects} as the proposal generator. It generates 2000 and 256 proposal boxes in training and testing, respectively. The RoI head is achieved by a 3-stage Cascade R-CNN~\cite{cai2018cascade}. The object box used by the text decoder is predicted from the last stage. The number of classes of each stage classifier is set to 2, \ie, foreground and background, where our objectness score is computed by averaging the foreground scores of all the three stages. For object detection task on COCO, soft NMS is applied during testing, and a mask head is added for multi-task training.

\myparagraph{Training.} In experiments, we explore two pre-training schemes: 1) MAE pre-training: The ViT backbone is initialized from the self-supervised MAE~\cite{he2022masked} trained on ImageNet-1K~\cite{deng2009imagenet}, while the rest of the model parameters are randomly set; 2) GIT pre-training: The ViT backbone and text decoder are initialized from the pre-trained image VL model GIT~\cite{wang2022git} and the rest are randomly set. We use MAE pre-training for most experiments by default unless otherwise specified. For the model initialized from GIT pre-training, we finetune on the downstream tasks by 90k iterations with a training batch size of 32. We find MAE pre-training can alleviate overfitting and benefit from more training epochs as discussed in~\cite{li2022exploring}. Thus, for the model initialized from MAE pre-training, we increase finetuning iterations to 180k and batch size to 64. We use the AdamW optimizer~\cite{loshchilov2017decoupled} with a learning rate of $8\times10^{-5}$ and the cosine learning rate decay schedule. In training, the input image size is $1024\times1024$ pixels resized by the large-scale jittering~\cite{ghiasi2021simple}. The testing image size is $800\times1300$ pixels.

\begin{figure*}[t!]
	\centering
	\includegraphics[width=1\linewidth]{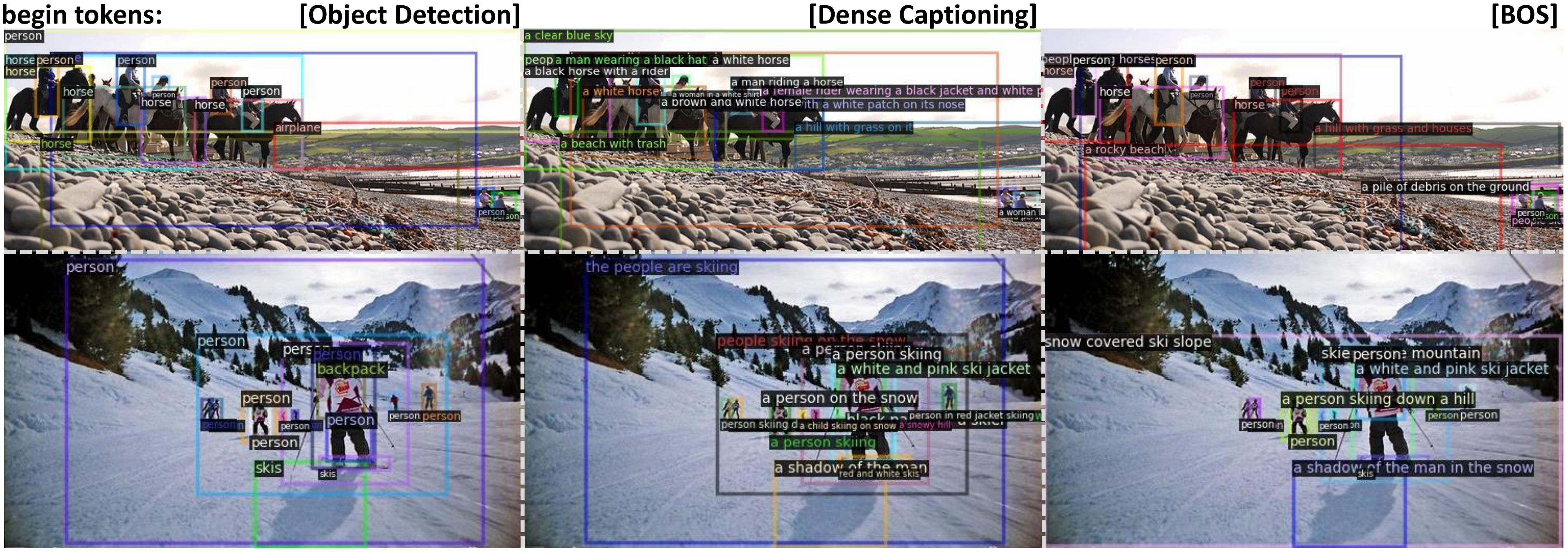}
	\vspace{-5mm}
	\caption{\textbf{Predictions from the models jointly trained on object detection and dense captioning.} Column \textcolor{red}{1\&2}: the model using two begin tokens \texttt{[ObjectDet]} and \texttt{[DenseCap]}. Column \textcolor{red}{3}: the model using only one begin token \texttt{[BOS]}. To generate object descriptions in a specific style, it is necessary to instruct the model with a unique begin token for that style. Zoom in for the best viewing}
	\label{fig:fig5}
\end{figure*}

\subsection{Joint Object Detection and Dense Captioning}
Since GRiT is a general object understanding model and can generate any style of object descriptions in the same framework, we jointly train a model on both object detection (short-description task) and dense captioning (long-description task). As shown in Table~\ref{tab:joint}, we compare the jointly trained model with two models that are trained on each task separately. We find that the separately trained model outperforms the jointly trained model. We hypothesize the main cause is that the COCO and VG datasets are not in consensus about box annotations. All boxes in COCO are located at foreground objects, while VG has lots of overlapped boxes and COCO-background boxes. COCO-background refers to the region that is not annotated in COCO. This leads to many false positives when testing COCO object detection as evidenced in the first column of Fig.~\ref{fig:fig5}. It may also cause low recall when testing VG dense captioning. We believe GRiT is capable of achieving both tasks in the same trained model without noticeable performance drops if the box data have no disagreement. In future work, we may study a dataset-specific foreground object extractor such that the extractor can be instructed to predict the boxes in certain dataset styles. In view of the dataset conflict, we train GRiT separately on these two datasets in the following experiment sections for more reliably studying GRiT's properties and fairly comparing with other single-task methods. 

As discussed in the method section, we instruct GRiT to generate task-specific descriptions by the begin tokens $\{$\texttt{[task]}$_i\}_{i=1}^{T}$ when jointly training the tasks with different styles of descriptions. To demonstrate this adaptation is the key to multi-task object understanding in one trained model, we compare to the model that is jointly trained on both tasks using only one begin token \texttt{[BOS]}. As shown in the last column of Fig.~\ref{fig:fig5}, the model with only \texttt{[BOS]} token cannot generate consistent object descriptions in the same image. Some objects are described in the way of dense captioning, while others are simply described by COCO class names. We also notice that COCO-background regions are more likely described by descriptive sentences because such regions only exist in the VG dataset during training. In contrast, the model informed by \texttt{[task]}$_i$ token correctly generates descriptions in the style we request.

\begin{table}[t!]
	\begin{center}	
		\setlength{\tabcolsep}{10pt}
		\begin{tabular}{c|cc}
			\rowcolor{mygray}
			&\multicolumn{2}{c}{{Testing Task}}\\
			\rowcolor{mygray}
			\multirow{-2}{*}{Training Task}&ObjectDet&DenseCap\\
			\shline
			ObjectDet, DenseCap&50.7&14.37\\
			ObjectDet&53.8&-\\
			DenseCap&-&15.48\\
			\hline
		\end{tabular}
	\end{center}
	\vspace{-3mm}
	\caption{\textbf{Jointly \emph{vs.} Separately training object detection and dense captioning.} Object detection and dense captioning are evaluated on COCO 2017 test-dev and VG test, respectively. All experiments are based on ViT-B.}
	\label{tab:joint}
\end{table}

\begin{table*}[t]
	\begin{subtable}[t]{0.33\linewidth}
		\centering
		\captionsetup{width=0.9\linewidth}
		\begin{tabular}{c|c|cc} 	
		\rowcolor{mygray}
		Method&AP&AP$_{50}$&AP$_{75}$\\
		\shline
		Closed-set OD&\textbf{51.7}&70.0&56.4\\
		GRiT&50.9&68.6&55.6\\
		\end{tabular}
		\caption{\textbf{GRiT vs. Closed-set object detector.} GRiT is comparable to closed-set standard object detector. Closed-set OD follows the same setting as GRiT but replaces the text decoder with a closed-set classifier as in standard OD.}
		\label{tab:ablation1}
	\end{subtable}
	\begin{subtable}[t]{0.33\linewidth}
		\centering
		\captionsetup{width=0.9\linewidth}
		\begin{tabular}{c|c|cc} 	
		\rowcolor{mygray}
		Size&AP&AP$_{50}$&AP$_{75}$\\
		\shline
		$7\times7$&50.8&68.4&55.6\\
		$14\times14$&\textbf{50.9}&68.6&55.6\\
		\end{tabular}
		\caption{\textbf{Object feature size.} GRiT is not sensitive to the number of object features.}
		\label{tab:ablation2}
	\end{subtable}
	\begin{subtable}[t]{0.33\linewidth}
	\small
		\centering
		\captionsetup{width=0.9\linewidth}
		\begin{tabular}{c|c|cc} 	
		\rowcolor{mygray}
		Beam size&AP&AR@1&AR@10\\
		\shline
		1&50.0&37.0&61.4\\
		2&50.8&38.0&63.6\\
		3&\textbf{50.9}&38.3&64.0\\
		5&\textbf{50.9}&\textbf{38.4}&\textbf{64.1}\\
		\end{tabular}
		\caption{\textbf{Beam search.} Beam search improves especially recall by labeling one box with more than one class name. Beam search is disabled when beam size = 1.}
		\label{tab:ablation3}
	\end{subtable}
		
	\begin{subtable}[t]{0.5\linewidth}
		\centering
		\captionsetup{width=0.9\linewidth}
		\setlength{\tabcolsep}{3pt}
		\begin{tabular}{c|c|c|c|cc} 	
		\rowcolor{mygray}
		Beam size&Objectness&Description&AP&AR@1&AR@10\\
		\shline
		\multirow{2}{*}{1}&\checkmark&&49.1&36.9&61.3\\
		&\checkmark&\checkmark&\textbf{50.0}&\textbf{37.0}&\textbf{61.4}\\
		\hline
		\multirow{2}{*}{3}&\checkmark&&19.7&32.3&61.3\\
		&\checkmark&\checkmark&\textbf{50.9}&\textbf{38.3}&\textbf{64.0}
		\end{tabular}
		\caption{\textbf{Object scoring.}  Object description score is crucial to remove false alarms when there is more than one label for each box.}
		\label{tab:ablation4}
	\end{subtable}
	\begin{subtable}[t]{0.5\linewidth}
		\centering
		\captionsetup{width=0.9\linewidth}
		\setlength{\tabcolsep}{3pt}
		\begin{tabular}{c|c|c|c|c} 	
		\rowcolor{mygray}
		Training progress&0-60k&60k-80k&80k-90k&AP\\
		\shline
		\multirow{3}{*}{\makecell{Training\\ object classes}}&60&60&80&49.1\\
		&60&80&80&50.4\\
		&80&80&80&\textbf{50.9}\\
		\end{tabular}
		\caption{\textbf{Incremental training.} GRiT seamlessly learns new object classes that are added in the middle of training.}
		\label{tab:ablation5}
	\end{subtable}
		
	\caption{\textbf{Ablation studies} on COCO 2017 val. All models are based on ViT-B and trained by 90k iterations with a batch size of 32.}\label{tab:ablation}
\end{table*}

\begin{table}[t!]
	\vspace{-2mm}
	\begin{center}	
		\setlength{\tabcolsep}{1pt}
		\begin{tabular}{ccc|c|c}
			\rowcolor{mygray}
			\multicolumn{3}{c|}{{\emph{Pre-training}}}&&\\
			\rowcolor{mygray}
			Method&Task (Data)&Parameters&\multirow{-2}{*}{Backbone}&\multirow{-2}{*}{AP}\\
			\shline
			\multirow{3}{*}{GIT~\cite{wang2022git}}&\multirow{3}{*}{\makecell{Language Modeling\\(Image-text pairs)}}&\multirow{3}{*}{\makecell{backbone,\\
			text decoder}}&ViT-B&52.0\\
			&&&ViT-L&52.7\\
			&&&Coswin-H&54.8\\
			\hline
			\multirow{3}{*}{MAE~\cite{he2022masked}}&\multirow{3}{*}{\makecell{Image\\Reconstruction\\(ImageNet-1K)}}&\multirow{3}{*}{\makecell{backbone}}&ViT-B&53.6\\
			&&&ViT-L&56.3\\
			&&&ViT-H&58.8\\
			\hline
		\end{tabular}
	\end{center}
	\vspace{-5mm}
	\caption{\textbf{GRiT pre-training.} MAE pre-training outperforms GIT pre-training, especially in the huge size. GIT adopts Coswin-H~\cite{yuan2021florence} as the visual backbone, so we adjust our backbone accordingly. Results are evaluated on COCO 2017 val.}
	\label{tab:pretrain}
	\vspace{-4mm}
\end{table}

\subsection{Ablation Studies}
In this section, we perform ablation experiments on object detection task to study GRiT's properties.

\myparagraph{GRiT \emph{vs.} Closed-set Object Detector:} GRiT achieves object detection in an \emph{open-set} way, which is more difficult than the closed-set way of standard object detectors. To measure the performance gap between these two strategies, we build a closed-set object detector by replacing GRiT's text decoder with the closed-set multi-category classifier as used in standard object detectors. The rest of the model settings are exactly the same as GRiT's settings. As shown in Table~\ref{tab:ablation1}, GRiT is comparable to the closed-set object detector with a 0.8 AP gap. This result demonstrates GRiT's open-set framework can serve as a new promising formulation for object detection.

\begin{table*}[t!]
	\vspace{-2mm}
	\begin{center}	
	\setlength{\tabcolsep}{4pt}
		\begin{tabular}{l|c|c|c|ccc|ccc}
			\rowcolor{mygray}
			&&&& \multicolumn{3}{c|}{{ 2017 val}} & \multicolumn{3}{c}{{2017 test-dev}}\\
			\rowcolor{mygray}
        \multirow{-2}{*}{Model} & \multirow{-2}{*}{Backbone}  &
        \multirow{-2}{*}{\makecell{Backbone\\Pre-training}}  &\multirow{-2}{*}{\makecell{Full model\\Pre-training}} & AP & AP$_{50}$ &AP$_{75}$& AP & AP$_{50}$ &AP$_{75}$\\
            \shline
            \multicolumn{3}{@{}l}{\emph{Closed-Set framework:}}\\
            \hline
           Deformable DETR~\cite{zhu2020deformable}&ResNeXt-101&IN-1K&-&-&-&-&50.1&69.7&54.6\\
           EfficientDet-D7x~\cite{tan2020efficientdet}&EfficientNet-B7&IN-1K&-&54.4&-&-&55.1&74.3&59.9\\ 
           CenterNet2~\cite{zhou2021probabilistic}& Res2Net&IN-1K& Unlabeled COCO&-&-&-&56.4&74.0&61.6\\
            HTC++~\cite{liu2021swin,chen2019hybrid}&Swin-L&IN-22K&-&57.1&-&-&57.7&-&-\\
            DyHead~\cite{dai2021dynamic}&Swin-L&IN-22K&-&58.4&-&-&58.7&77.1&64.5\\
            ViT-Adapter-L~\cite{chen2022vision}&ViT-L&IN-22K&-&58.4&-&-&58.9&-&-\\
             Soft Teacher~\cite{xu2021end}&Swin-L&IN-22K&Object365&60.1&-&-&-&-&-\\
            ViTDet~\cite{li2022exploring}&ViT-H&IN-1K&-&60.4&-&-&-&-&-\\
            DINO~\cite{zhang2022dino}&Swin-L&IN-22K&Object365&63.1&-&-&63.2&-&-\\
            GLIPv2~\cite{zhang2022glipv2}&Swin-H&\makecell{CC-15M+\\IN-22K}&\makecell{Object365, FourODs,\\GoldG, CC15M+SBU}&-&-&-&60.6&-&-\\
			\shline
			\multicolumn{3}{@{}l}{\emph{Open-Set framework:}}\\
			\hline
			GRiT (Ours)&ViT-B&IN-1K&-&53.6&71.6&58.2&53.8&71.8&58.7\\
			GRiT (Ours)&ViT-L&IN-1K&-&56.3&73.8&61.4&56.6&74.5&61.8\\
			GRiT (Ours)&ViT-H&IN-1K&-&58.8&76.6&64.4&59.0&76.9&64.5\\
			GRiT (Ours)&ViT-H&IN-1K&Object365&60.3&78.1&65.7&60.4&78.1&66.0\\
			\shline
		\end{tabular}
	\end{center}
	\vspace{-5mm}
	\caption{\textbf{Comparison with state-of-the-art object detectors on COCO.} All results are reported under single-scale testing. The reference results are cited from the best-performing models in their papers. }
	\label{tab:sota_od}
	\vspace{-1mm}
\end{table*}

\myparagraph{Object Feature Size:} We experiment with different sizes of object features input to the text decoder. As shown in Table~\ref{tab:ablation2}, 49 feature vectors achieve similar performance to 196 feature vectors, which indicates the text decoder is robust to the number of input object features.

\myparagraph{Beam Search:} Standard object detector outputs multiple class labels for one box by its multi-category classifier. Similarly, we use beam search to output multiple class name texts for each box as introduced in the method section. As shown in Table~\ref{tab:ablation3}, beam search improves object detection metric, especially for recall, and beam size=3 is a good trade-off between accuracy and inference time.

\myparagraph{Object Scoring:} To rate object predictions, we combine both objectness score from the foreground object extractor and object description score from the text decoder. GRiT is always equipped with objectness score due to its function of removing background boxes. As shown in Table~\ref{tab:ablation4}, description score improves 0.9 AP when beam size=1. However, the model without description score fails when beam size=3, \ie, outputting three class names each box. The reason is that all the three classes share the same confidence score though at least two of them are false positives. This has a mild impact on recall but leads to significantly worse precision and AP.

\myparagraph{Incremental Training:} GRiT is open-set and capable of generating unlimited number of words. As data evolve, one can add new object classes or concepts in the middle of training without adapting any architecture. We simulate this use case on COCO in Table~\ref{tab:ablation5}, where we start training with 60 classes and add the remaining 20 classes in the middle of training. Compared to the model that is trained on all classes throughout, we achieve similar results when adding the rest of the classes in the last one-third of training. GRiT performs reasonably even in the case where we supplement the remaining classes in the last one-ninth of training.

\begin{figure*}[t!]
	\vspace{-1mm}
	\centering
	\includegraphics[width=1\linewidth]{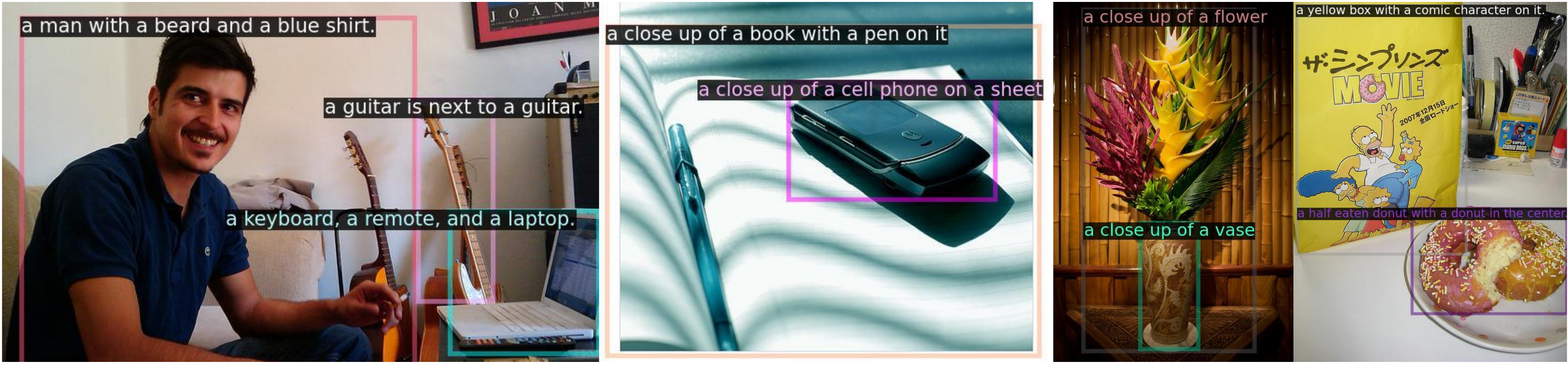}
	\vspace{-5mm}
	\caption{\textbf{Zero-shot object understanding predictions.} Zoom in for the best viewing.}
	\vspace{-2mm}
	\label{fig:fig6}
\end{figure*}

\myparagraph{Pre-training:} We study GRiT with different pre-training schemes. As shown in Table~\ref{tab:pretrain}, the performance improves notably as the model size increases, and MAE pre-training shows better performance than GIT pre-training. MAE pre-training is to recover the masked patches in the image, which may exhibit a stronger localization capability helping object detection task. GIT pre-training focuses more on the image-level representation and language modeling, which shows slightly better performance on dense captioning task, as in Table~\ref{tab:sota_densecap}.

\subsection{Comparison to State-of-the-Arts on COCO}
As shown in Table~\ref{tab:sota_od}, we compare GRiT with the state-of-the-art object detectors on COCO. Generally, all the methods use the visual backbone pre-trained on ImageNet (IN). To deliver the best performance, some models are also finetuned on extra object detection datasets, \eg, Object365~\cite{shao2019objects365}, before finetuning on COCO. The results of the state-of-the-art object detectors are cited from their best model settings. Therefore, some listed models may be achieved in different training and testing configurations than others. For example, GLIPv2~\cite{zhang2022glipv2} makes use of Object365 plus four object detection datasets~\cite{zhang2022glipv2}, image-text datasets CC~\cite{sharma2018conceptual} and SBU~\cite{ordonez2011im2text}, and grounding datasets GoldG~\cite{zhang2022glipv2}. DyHead~\cite{dai2021dynamic} utilizes a larger input image size with 2000 pixels at maximum. CenterNet2~\cite{zhou2021probabilistic} adopts BiFPN~\cite{tan2020efficientdet}, DCN~\cite{dai2017deformable}, and a larger input image size of 1560$\times$1560 pixels.  Soft Teacher~\cite{xu2021end} is constructed on HTC++~\cite{liu2021swin}. Despite the challenge of the open-set framework, GRiT performs comparably with the state-of-the-art closed-set object detectors. This once again demonstrates our generative region-to-text framework is a promising formulation for excelling in object detection. 

\subsection{Comparison to State-of-the-Arts on VG}
As shown in Table~\ref{tab:sota_densecap}, we compare GRiT with the state-of-the-art dense captioning models. Without object relation modeling as used in previous dense captioning methods, GRiT outperforms the current best model by 4 mAP. We notice that the absolute values are low on the dense captioning metric. We hypothesize the cause is the very dense region annotations and the strictness of language metric.

\begin{table}[t]
\vspace{-3mm}
	\begin{center}	
		\setlength{\tabcolsep}{10pt}
		\begin{tabular}{l|c}
			\rowcolor{mygray}
			Method&mAP\\ 
			\shline
            FCLN~\cite{johnson2016densecap}&5.39\\
            JIVC~\cite{yang2017dense}&9.31\\
            ImgG~\cite{li2019learning}&9.25\\
            COCD~\cite{li2019learning}&9.36\\
            COCG~\cite{li2019learning}&9.82\\
            CAG-Net~\cite{yin2019context}&10.51\\
            TDC+ROCSU~\cite{shao2022region}&11.49\\
            \hline
            GRiT$_\text{MAE}$ (Ours)&15.48\\
			GRiT$_\text{GIT}$ (Ours)&15.52\\
			\hline
		\end{tabular}
	\end{center}
    \vspace{-4mm}
	\caption{\textbf{Comparison with state-of-the-art dense captioning models on Visual Genome.} GRiT$_\text{MAE}$ refers to the model initialized by MAE pre-training scheme. GRiT$_\text{GIT}$ is initialized from the GIT model that is re-pretrained on CC3M and CC12M~\cite{sharma2018conceptual} datasets removing the VG dataset. Our results are reported based on ViT-B.}
	\vspace{-4mm}
	\label{tab:sota_densecap}
\end{table}

\subsection{Zero-shot Object Understanding}
Inspired by the success of image-to-text transformers on image understanding, GRiT adopts the same network structure for the visual encoder backbone and text decoder as GIT~\cite{wang2022git}. We explore whether GRiT can achieve zero-shot object understanding by simply using GIT's trained parameters without finetuning on object description annotations. To this end, we initialize GRiT by the GIT model fine-tuned on COCO image captioning task.
Then, we finetune our feature pyramid and foreground object extractor on COCO detection data (no use of the class names) while keeping GIT's parameters fixed, such that GRiT is able to generate object boxes. To align with GIT's parameters, object features are cropped from the last feature map of the visual backbone rather than the feature pyramid. This zero-shot object understanding result is shown in Fig.~\ref{fig:fig6}. We see that the model generates various object-level descriptions for different regions in the same image though GIT is trained on the image-level description task. While, we also notice in experiments that there are images for which this zero-shot scheme does not work well. We hypothesize the issue is that GIT's text decoder parameters do not well recognize the cropped features as GIT's text decoder is trained with the entire image's features as input. 

\section{Conclusion}
This work presents a general and open-set object understanding framework, GRiT. GRiT formulates object understanding as region-text pairs, which is capable of unifying various region/object-level tasks in a single paradigm. GRiT is end-to-end from image feature extraction to foreground object detection to object description generation. Extensive experiments on object detection and dense captioning demonstrate the effectiveness and generality of GRiT. 

{\small
\bibliographystyle{ieee_fullname}
\bibliography{GRiT}
}
\end{document}